%% file: main.tex
\documentclass[conference,10pt]{IEEEtran}

    \IEEEoverridecommandlockouts
    \usepackage{pifont}
    \usepackage{booktabs} % For formal tables
    \usepackage{footnote}
    \usepackage[pass]{geometry}
    \usepackage[english]{babel}
    \usepackage{comment}
    \usepackage[linesnumbered,algoruled]{algorithm2e}
    \usepackage{amsmath}
    \usepackage{bm}
    \usepackage{graphicx}
    \usepackage{multicol}
    \usepackage{amssymb}
    \usepackage{setspace}
    \usepackage{multirow}
    
\begin{document}
\sloppy
      
\title{CoopNet: Cooperative Convolutional Neural Network for Low-Power MCUs\vspace*{-2mm}}
      
\author{Luca Mocerino, Andrea Calimera\\
    Politecnico di Torino, 10129, Torino Italy\vspace*{-2mm}}
    
%\IEEEpubid{\makebox{\hfill \normalsize 978-1-5386-4756-1/18/\$31.00~\copyright~2018 IEEE}
%\hspace{\columnsep}\makebox[\columnwidth]{}}
    
%\IEEEpubid{\makebox[\columnwidth]{978-1-5386-4756-1/18/\$31.00~\copyright2018 IEEE \hfill} \hspace{\columnsep}\makebox[\columnwidth]{ }}
    
\maketitle
%\IEEEpubidadjcol
    
\begin{spacing}{0.93} %91
    \begin{abstract}
Fixed-point quantization and binarization are two reduction methods adopted to deploy Convolutional Neural Networks (CNNs) on end-nodes powered by low-power micro-controller units (MCUs). While most of the existing works use them as stand-alone optimizations, this work aims at demonstrating there is margin for a joint cooperation that leads to inferential engines with lower latency and higher accuracy.
% that are higher than those achieved when the two techniques are run as separated. 
Called {\em CoopNet}, the proposed heterogeneous model is conceived, implemented and tested on off-the-shelf MCUs with small on-chip memory and few computational resources. Experimental results conducted on three different CNNs using as test-bench the low-power RISC core of the Cortex-M family by ARM validate the CoopNet proposal by showing substantial improvements w.r.t. designs where quantization and binarization are applied separately.
    \end{abstract}
        
   \section{Introduction}
    \input{intro}
          
    \section{Background}\label{sec:back}
    \input{background}

    \section{CoopNet}
    \input{work}

    \section{Experimental results}
    \input{results}
    %\vspace{-0.8cm}
    \section{Related works}
    Fixed-point quantization \cite{fixquant}, as well as  binarization \cite{BNN,BinCon,xnor}, represent a valuable solution to deploy CNN on ultra-low-power commercial MCUs. While 8-bit are almost sufficient to guarantee the same accuracy of floating-point models, lower bit-widths lead to inaccurate inference. Recent works investigated on arbitrary bit-widths (i.e. $2\leq bit<8$). More specifically, they aim at finding the optimal balance between accuracy, resource utilization and performances assigning different bit-widths to different layers \cite{mixp}. Unfortunately, commercial low-power MCUs do not have programmable data-paths and memory interfaces to support arbitrary bit-width arithmetic efficiently \cite{rusci2018}. 
    Amiri et al. in \cite{het} proposed a system level mixed-precision solution which exploits heterogeneous CPU and FPGA accelerators. The overhead, both on-line (due to a tuning procedure) and off-line (during training), and the resources required make this approach less suitable for low-end MCUs. Combining multiple CNN models into {\em ensemble} results in a winning solution for many tasks \cite{hydra}.
    However, the resources required to host several models and execute them in parallel make this approach practically not scalable on a low-end device.
    On the contrary, CoopNet enables an efficient and accurate solution for off-the-shelf MCUs proposing a flexible architecture adaptable to the user-defined constraint.

\begin{table}[t]
 \resizebox{\columnwidth}{!}{
    \centering
   
    \begin{tabular}{|c|c|c|c|c|c|}
        \hline 
        \textbf{\shortstack{Dataset\\(Net)}}&\textbf{\shortstack{Accuracy \\Level}}& \shortstack{$\mathbf{\Delta}$\\ {\bf(\%)}}&\textbf{\shortstack{Mem. Size\\ (kB)}}& \textbf{\shortstack{Speed-up \\(\%)}} &\textbf{ST}  \\ \hline \hline  \rule{0pt}{2ex}  
        \multirow{3}{*}{\shortstack{CIFAR-10 \\(CaffeNet)}}&FP32&+0.05 & &51.20 &0.42  \\
         &INT8&0 &214 & 51.58 &0.4\\ 
          &Max-accuracy &+0.53& & 18.40  &0.9 \\ \hline 
         \multirow{3}{*}{\shortstack{GSC \\(GscNet)}}&FP32 &+0.8&& 69.53 &0.4 \\
         &INT8&0 &360&80.16 &0.2\\ 
         &Max-accuracy &+1.47& & 54.31 &0.7 \\ \hline\rule{0pt}{2ex}
         
           \multirow{3}{*}{\shortstack{FER13 \\(FerNet)}}&FP32 &+0.46&& 36.20 &0.3 \\ 
         &INT8&0&695& 47.90 &0.2  \\ 
          &Max-accuracy &+0.93&& 22.34  &0.5 \\ \hline
    \end{tabular}}
    \vspace{3mm}
    \caption{CoopNet: Main achievements and Final Results}
    
    \label{tab:fin_res}
    %delta in the table
\end{table}
    
    \section{Conclusions}
    CoopNet is a novel network architecture that integrates a fast and unreliable model with a slower but accurate one to improve the processing efficiency of inference models. The joint cooperation of binary and 8-bit quantized models guarantees higher accuracy and substantial speed-up, also offering a valuable option for adaptive energy-accuracy inference on the edge. 
     %\vspace{-0.1cm}
     
\end{spacing}

\bibliographystyle{IEEEtran}
\bibliography{refs}
\end{document}

%% file: intro.tex
Inference engines built upon end-to-end deep learning methods represent the state-of-the-art in several application domains. Deep Convolutional Neural Networks (CNNs), in particular, have brought about breakthroughs in the field of computer vision, speech recognition and natural language processing \cite{cnn_general}. Many Internet-of-Things (IoT) services rely on CNNs to infer information from the raw data gathered by end-user portable devices and/or embedded sensors. While the majority of IoT frameworks run CNNs in the cloud, namely, on centralized data centers physically very far from the source of data, to have CNNs on hand is a means to higher efficiency and more user privacy \cite{shi2016edge}. Enabling the inferential stage on the mobile edge is challenging as it requires the processing of CNNs, large in size and computationally intensive, with limited hardware resources. The picture gets even more complicated when considering applications, like wearable \cite{cnn_general} or ambient and infrastructural sensors \cite{electronics7100222}, which must run on tiny cores with few hundreds of kByte of on-chip memory and an active power consumption below the 100 mW mark. As practical example, this work considers the micro-controller units (MCUs) of the Cortex-M family designed by ARM for the IoT segment\footnote{https://os.mbed.com/platforms}.
%where the main concern is to reduce integration costs, improve the form-factor and minimize the energy budget.%  to enable the use of low-capacity batteries.
%In such cases the only option available for designers is to shrink the complexity of the CNN models till they can fit tiny cores with limited resources: few hundreds of KByte of on-chip memory and an active power consumption below a the 100 mW mark, such as for instance the microcontroller units (MCUs) of the Cortex-M family designed by ARM for the IoT \footnote{https://os.mbed.com/platforms}.
In such cases, the only available option is to shrink down the cardinality of the CNN model until it fits the underlying hardware architecture.

Among the available algorithmic optimizations, post-training quantization via integer arithmetic has become a must-do stage: most of the MCU cores deployed on the end-nodes do not have floating-point units indeed. The use of arithmetic representations with scaled bit-widths helps to reduce the memory footprint, but above all it ensures a larger memory bandwidth as multiple data can be packed within the same word. This ensures lower latency and hence smaller energy consumption w.r.t 32-bit floating point. In \cite{fixquant} the authors demonstrate that 8-bit fixed-point integer guarantees near-to-zero accuracy loss with 4$\times$ memory reduction. Extreme quantization to 1-bit \cite{BinCon,BNN,xnor} leads to Binary CNNs with the smallest footprint, but also the lightest workload as (some) integer arithmetic get replaced with bit-wise Boolean operators. 
%Needless to say, binary CNNs are very attractive for running inference at the edge. 
However, binary CNNs come with significant accuracy loss: from 2\%, up to 10\%, 20\%, and even more, depending on the original CNN and the complexity of the training data-set. This represents a key limiting factor.
%their use in practical applications

This work aims to address this drawback demonstrating there exist margins to exploit binary CNNs for building highly accurate, yet fast inferential models that can be deployed on the edge. The proposed solution, called {\em Cooperative Convolutional Neural Network} (CoopNet), consists of a joint combination of binary and 8-bit fixed-point CNN models controlled by a probabilistic thresholding policy. The resulting heterogeneous inferential model improves the classification accuracy and meets the hardware constraints of low-power MCUs. Experimental results, conducted on three CNNs trained to run classification on three data-set belonging to different domains, reveal CoopNet deployed on the Cortex-M cores by ARM outperforms classical homogeneous CNN, therefore achieving an improved accuracy-latency tradeoff.

\begin{comment}
Deep Learning algorithms (DL) and in particular Convolutional Neural Networks (CNNs)  have become the standard {\em de facto} for several computer vision tasks such as object detection, recognition \cite{} and image segmentation \cite{}. However, the intensive computational requirements and the huge memory footprint make the CNNs mapping on low-power edge devices challenging. Therefore, the cutting-edge research devotes a large number of efforts in aggressive algorithmic optimization to deploy such CNNs into edge-devices. The recent literature found in the parameters quantization a suitable solution for reducing the algorithmic complexity. In particular, Lin et. al in \cite{fixquant} demonstrated that with 8-bit fixed-point quantization the accuracy drop is near-to-zero with respect to the floating point. Meanwhile, the extreme quantization approaches for Binarized Neural Networks (BNNs) proposed in \cite{BinCon,BNN,xnor} revealed that it is possible to drastically reduce the memory footprint and the computational data-path at the cost of significant accuracy loss (from 3 to 10 \%).
\end{comment}

\begin{comment}
\begin{table}[ht]
    \centering
    \begin{tabular}{c|c|c|c|}
        & \textbf{Accuracy} & \textbf{Latency} \\ \hline
         q-bit  & \ding{51} & \ding{55} \\
         BNN & \ding{55} & \ding{51} \\
         CoopNet & \ding{51} & \ding{51} \\ 
    \end{tabular}
    \vspace{2mm}
    \caption{Caption}
    \label{tab:my_label}
\end{table}
\end{comment}

%% file: background.tex
\begin{figure}[!t]
\centering
    \includegraphics[scale=0.53]{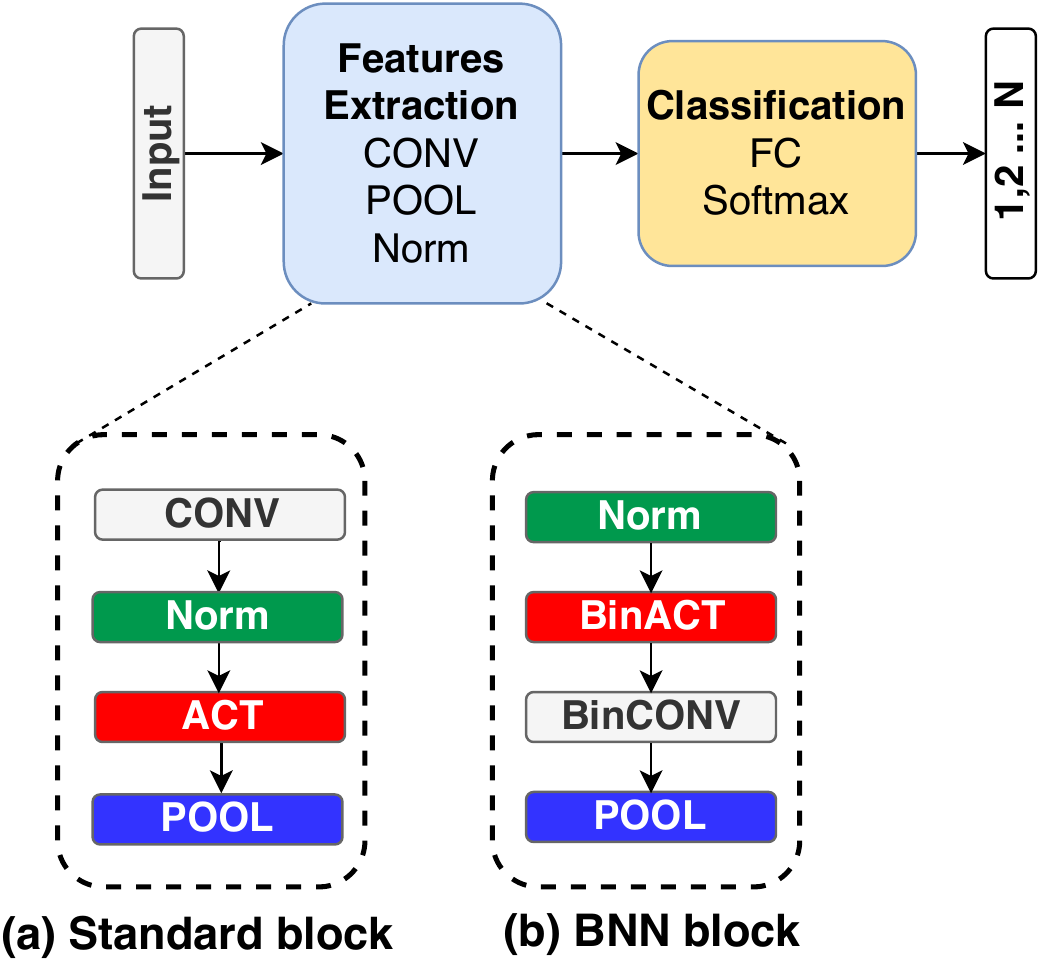}
    \caption{Structure of a typical CNN. (a) Basic block of a standard Integer CNN; (b) Basic block of a BNN with binarized weights and activations}
    \label{fig:simple_cnn}
\end{figure}

\subsection{Convolutional Neural Networks (CNNs)}
CNNs are a special class of end-to-end trainable models mostly suited for the classification of multi-dimensional spatial inputs, like multi-channel images. They consist of several computational layers chained to form a deep architecture. Existing CNNs mainly differ for their internal topology, namely, how different kinds of hidden layers are sized and connected. It is however possible to recognize a common structure which is made up of two macroblocks (Fig.~\ref{fig:simple_cnn}): {\em Feature Extraction}, where relevant features learned during the training stage are extracted layer-wise using kernel convolutions; {\em Classification}, where the extracted features get classified.

Within the feature extraction block, the most commonly adopted layers are: {\em convolutional} layers (CONV), which perform multidimensional convolutions between the output tensor generated by the previous layer (also called feature map) and local filter tensors; {\em pooling} layers (POOL), e.g. max pooling or average pooling, which reduce the dimension of feature maps; {\em normalization} layers (Norm), that normalize the distribution (mean and standard deviation) of the activation maps; {\em activation} function (ACT), e.g. ReLU or tanh, which introduces non-linearity. The classification block is built upon {\em fully-connected} layers (FC), which implement a geometric separation of the extracted features, and {\em softmax}, that produces a probability distribution over the available classes.
%{\em dropout} layers that help to mitigate over-fitting 

\begin{comment}
Such layers can be grouped in three main regions, each with a specific function: {\em input layers} handle inputs for computational stages, {\em hidden} layers extract relevant features learned during the training stage, {\em output layers} take the extracted features and run the actual classification that brings to the final labeling. Existing CNNs mainly differ for the topology of the feature extraction region and the classification region. The former is built upon different kinds of layers arranged following different topology. There might be {\em convolutional} (CONV) layers that perform multidimensional convolutions between the output generated by the previous layer and some local filter, {\em pooling} layers which reduce the dimension of feature maps, {\em dropout} layers that help to mitigate over-fitting. The latter implements a geometric separation of the extracted features and it typically consists of {\em fully-connected} (FC) neural network that computes scores on each feature map.
\end{comment}

\subsection{Fixed-Point Quantization}
While a CNN training is usually run using a 32-bits floating-point representation, recent studies, e.g. \cite{fixquant,hubara2017quantized}, demonstrate that fixed-point integers with lower arithmetic precision are enough for inference. Fixed-point quantization is becoming a consolidated standard when the target hardware are low-power cores with small memory footprint and reduced instruction set (8/16-bit integer).
%This allows reducing the memory footprint and the inference computational cost exploiting a reduced precision, integer-only arithmetic.
%The literature is plenty of fixed-point quantization schemes, e.g. {\em static/dynamic}, {\em linear/non-linear}. 
A detailed review of all the quantization schemes in literature is out of the scope of this work and interested readers may refer to \cite{fixquant,hubara2017quantized,mixp}. This work adopts the $q$-bit fixed-point quantization proposed in \cite{hubara2017quantized}. The convolution run in a CONV layer between the input feature map ${\bf x} \in \mathbb{R}^{c \times w_{in}\times h_{in}}$ and the local weights ${\bf w} \in \mathbb{R}^{c \times kw\times kh} $ is as follows:
\begin{equation}
    \textbf{x} \ast \textbf{w} = 2^{-2(q-1)} \sum_{i \in C} X_i \cdot W_i
\end{equation}
with $C$ as the number of channels. We set $q=8$ for both weights and activations, and $q=16$ for intermediate results accumulation. 

%We exploit that quantization schema for the baseline model {\em INT8} model. {\bf ANDREA: not clear}

\subsection{Binarized Neural Networks}
Several works proposed CNNs with binary weights and/or activations. BinaryConnect \cite{BinCon} represents the ancestor: weights are binarized using \textit{hard sigmoid} function, while activations remain in full-precision to avoid accuracy drop. The Binarized Neural Networks proposed in \cite{BNN} are the first example of fully binary CNN: both weights and activations are binarized via \textit{sign} function. The CONV layers are simplified through bit-wise XNOR and bit-count. This allows to achieve the highest compression ($\sim32\times$), yet with substantial accuracy loss (up to 28.7\%).
The authors of XNOR-Net \cite{xnor} addressed this drawback introducing a new topology where the binary output of each CONV layer is first re-scaled through a full-precision Norm layer. Fig.~\ref{fig:simple_cnn}-b gives a pictorial description of the basic block deployed in the XNOR-Net, where the suffix {\em Bin} highlights binarized layers. 

The mathematical description of a binary convolution is as follows. 
Given $\textbf{x} \in \mathbb{R}^{c \times w_{in}\times h_{in}}$ as the input feature and $\textbf{w} \in \mathbb{R}^{c \times kw\times kh} $ as the weight tensor, their convolution is approximated as follows:
\begin{equation}
 \textbf{x} \ast \textbf{w} \approx \text{popcount}\;(\textbf{X}\;  \text{xnor}\;  \textbf{W}) \cdot K \cdot \alpha
\end{equation}
where $K$ and $\alpha$ are scaling factors. While weights (\textbf{W}) are binarized with the {\em sign} function only, the activations are first normalized and then binarized. These stages can be fused into a single layer that includes all batch normalization parameters: variance $\sigma^2$, mean $\mu$, scale $\gamma$, shift factor $\beta$, and $\epsilon$ for numerical stability. A feature map {\bf x} is binarized as follows: 
\begin{equation}
    \text{\bf X} = BinACT_{0,1} (x) = \begin{cases}
                \text{1 }  x \ge c \\
                 \text{0 } x < c \\
                \end{cases}
\end{equation}
where $c= \mu-\beta/\gamma \sqrt{\sigma^2 + \epsilon}$ is constant at inference time. We do not use $K$, that is the activations scaling factor, for computational efficiency reasons. We represent $c$ and $\alpha$ with 8-bit integers.

An efficient processing of XNOR-Net requires data-paths capable of performing bit-wise xnor, bit-counter and comparison. These operators can be implemented with specialized units in case of custom hardware \cite{finn}, or through software routines compiled using the instruction-set available on the target general purpose core \cite{rusci2018}.

%% file: work.tex
\subsection{Concept and Architecture}
\begin{figure}[!t]
\centering
    \includegraphics[scale=0.35]{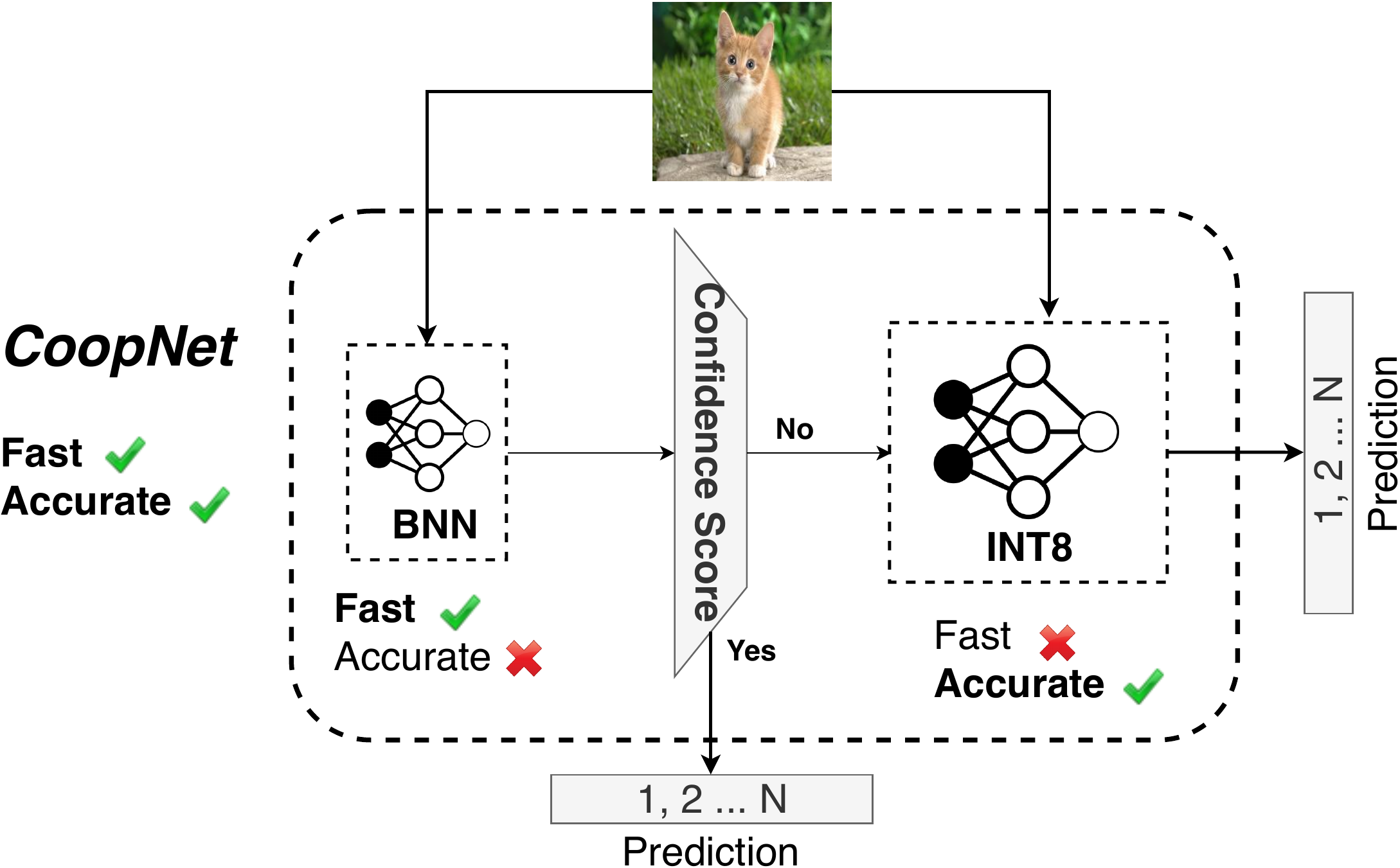}
    \caption{CoopNet concept and abstract architecture.}
    \label{fig:main_idea}
\end{figure}

The CoopNet inference concept is intuitive, yet very efficient. As graphically depicted in Fig. \ref{fig:main_idea}, it is based on the cooperation of two convolutional models: a binary net  BNN, fast and small but less accurate, an integer net INT8, slower and larger but more accurate.
The BNN processes the input data first. Then, if the prediction satisfies a certain criterion of confidence it is forwarded to the output as the outcome, otherwise the input is re-processed by the INT8 to produce a more confident output score. The criterion used to control the execution flow is called {\em Confidence Score (CS)} and it is defined as follows: 
\begin{equation}
    CS = P_{BNN}(y_{i}|x) - P_{BNN}(y_{j}|x)
\end{equation}
where $P_{BNN}(y_{n}|x)$ is the probability produced by the BNN  that a given input $x$ belongs to a given class $n \in 1,2\dots N $; $i$ and $j$ refer to the indexes of the first and second highest scored classes. 
Intuitively, a high {\em CS} means the BNN was able to classify the given input with enough confidence, on the contrary, a low {\em CS} means the topmost scored classes get very close to each other, which reveals a certain level of uncertainty, as the BNN was not able to make a clear distinction among the available classes. For the latter case, the INT8 model is activated for aid. The thresholding policy is controlled through a {\em Confidence  Threshold (CT)} which might be changed dynamically for run-time adjustments.

%\subsection{Design and Parameter Settings}
For a given task and application, the pre-trained 32-bit Floating Point model is used as basis to generate the INT8 model, obtained with the quantization method introduced in \cite{hubara2017quantized} using $q$=8-bit. The BNN model is built using the XNOR-Net method presented in \cite{xnor}. It is worth emphasizing that, according to \cite{xnor}, the first and last layers of the BNN model are kept to 8-bit.

A key design aspect concerns the setting of the threshold $CT$ as it affects the accuracy-latency trade-off. The parametric analysis reported in the experimental section provides a proper understanding of this important relationship.

\subsection{Extra-functional Metrics}
{\bf Latency.} Given a generic CoopNet, its latency is modeled through the following equation:
\begin{equation}
    L_{CoopNet}(CT) = \begin{cases}
               L_{BNN} + L_{CS}               & CS \geq CT\\
               L_{BNN} + L_{CS} + L_{INT8}    & CS < CT\\
           \end{cases}
           \label{eq:coop_latency}
\end{equation}
$L_{BNN}$ and $L_{INT8}$ are the latency of the BNN and INT8 models respectively, while $L_{CS}$ is the contribution due to $CS$ computation and comparison with $CT$. To notice that $L_{CS}$ is the latency of a single integer subtraction and comparison, hence its contribution is negligible w.r.t. the latency of BNN and INT. Both $L_{BNN}$ and $L_{INT8}$ can be simply estimated as the sum of the latency of each internal layer collected from on-board measurements. The layers characterization has been implemented using an extended version of the CMSIS-NN library by ARM \cite{cmsis} which supports binary convolutions \cite{rusci2018}. When considering batch inference, Equation \ref{eq:coop_latency} can be generalized as:
%\begin{equation}\label{eq:run_latency}
%    L_{BS} = \sigma\cdot L_{BNN} + (BS-\sigma) \cdot (L_{BNN}+ L_{INT}) 
%\end{equation}
\begin{equation}\label{eq:run_latency}
    L_{CoopNet}(BS, CT) = \sum_{i=1}^{BS} L_i(CT)
\end{equation}
where {\em BS} is the cardinality of the batch and $L_i(CT)$ is the latency of the $i$-th batch sample. 

%%% ANDREA: da inserire nella parte exp
%The Finally, we can compute the speed-up factor comparing the CoopNet latency in Eq \ref{eq:run_latency} with the overall latency when considering the baseline model only ($L_{baseline} = TS \cdot L_{INT}$ ). 

{\bf On-chip Memory.} The hardware cores targeted by this work are the smallest low-power MCUs equipped of the Cortex-M family by ARM. These MCUs are usually equipped with limited RAM ($\leq$ 1 MByte). The memory footprint of CoopNet is the sum of the RAM taken by the BNN model ($M_{bnn}$) and INT8 model ($M_{int}$). The two contributions include the RAM taken by the weights buffer, the activations buffer and {\em im2col} buffer as the model provided by ARM in \cite{cmsis}. The $CT$ parameter is one Byte, therefore negligible.

%% file: results.tex
\subsection{Experimental Setup}
CoopNet has been evaluated on the following three tasks:
\textbf{CIFAR-10 -} Image classification task; it consists of 60k $32\times32$ RGB images classified with 10 labels.\\
%(made up of 50k samples for training and 10k for testing)
\textbf{Google Speech Command  (GSC) - } Keyword spotting from speech; the data set \cite{gsc} collects 65k one-second long samples classified with 30 classes.\\ %The train set includes 56k samples while 9k for test set.
\textbf{Facial Expression Recognition (FER13) - } Emotion recognition from facial expression; the data set \cite{fer13} is made up of 36k $48 \times 48$ grayscale facial images classified by $7$ labels. %We adopt 29k samples for training and $7$k for testing.

Different lightweight CNNs suited for tiny cores are deployed for the three tasks. An overview is reported in Table \ref{tab:bench}; for sake of space, the table reports the CONV and FC layers together with their size, although there are activation, pooling and regularization (normalization and dropout) layers. Moreover, Table \ref{tab:bench} shows the top-1 accuracy (\%) and the memory footprint (kB) for full-precision (FP32), 8-bit fixed-point (INT8) and binary (BNN) models.

%The three CNNs are trained for $350$-epochs in PyTorch (version 1.0) using Adam optimization (learning rate $10^{-3}$, linear decay 0.1 every $100$-epochs, batch-size 128). 
%8-bit quantization and binarization from 32-bit floating point are the same introduced in  \cite{hubara2017quantized} and \cite{xnor} respectively.  

\begin{savenotes}
\begin{table}[!t]
\resizebox{\columnwidth}{!}{
    \centering
    \begin{tabular}{c|c||c|c|c|} \cline{2-5}
    &\multicolumn{1}{|c||}{\multirow{1}{*}{\textbf{Dataset}}}  & CIFAR-10  & GSC  & FER13 \\  \cline{2-5}
    &\multicolumn{1}{|c||}{\textbf{Model}}  & CaffeNet\footnote{Inspired by https://code.google.com/archive/p/cuda-convnet/} & GscNet \cite{gsc} & FerNet \\ \hline    
    \multicolumn{1}{|c|}{\multirow{2}{*}{\bf FP32}} &Accuracy (\%) &80.25 & 90.30  &65.16\\ 
     \multicolumn{1}{|c|}{} &Mem. Size (kB)&550   &1060    &2345 \\ \hline
    \multicolumn{1}{|c|}{\multirow{2}{*}{\textbf{INT8}}} & Accuracy (\%)  & 80.20 & 89.50 & 64.70  \\ 
     \multicolumn{1}{|c|}{}&Mem. Size (kB)  &120&250 &577 \\ \hline
    \multicolumn{1}{|c|}{\multirow{2}{*}{\textbf{BNN}}} &Accuracy (\%)  & 76.52 & 87.60 & 62.86  \\
     \multicolumn{1}{|c|}{}&Mem. Size (kB)  &94 &90 &118 \\ \hline \rule{0pt}{3ex}    
   &input & 3x32x32 & 1x32x32 & 1x44x44 \\ 
    && CONV 32x5x5 & CONV 32x5x5 & 3$\times$ CONV 32x3x3 \\
    &&  32x5x5 &  32x5x5 & 3$\times$ CONV 64x3x3\\
    &&  64x5x5&  64x5x5 &  3$\times$ CONV 128x3x3\\
     && FC 1024x10 & 64x5x5  &  FC 128x7 \\ 
    && &FC 1024x31 &  \\ \cline{2-5}
         
    \end{tabular}

    }
    \vspace{3mm}  
    \caption{Benchmarks: Datasets and CNNs\vspace{-3mm}}
    \label{tab:bench}
    
\end{table}
\end{savenotes}

\subsection{Performance Assessment}
\begin{figure}[ht]
\vspace*{-1mm}
    \centering
    \includegraphics[scale=0.43]{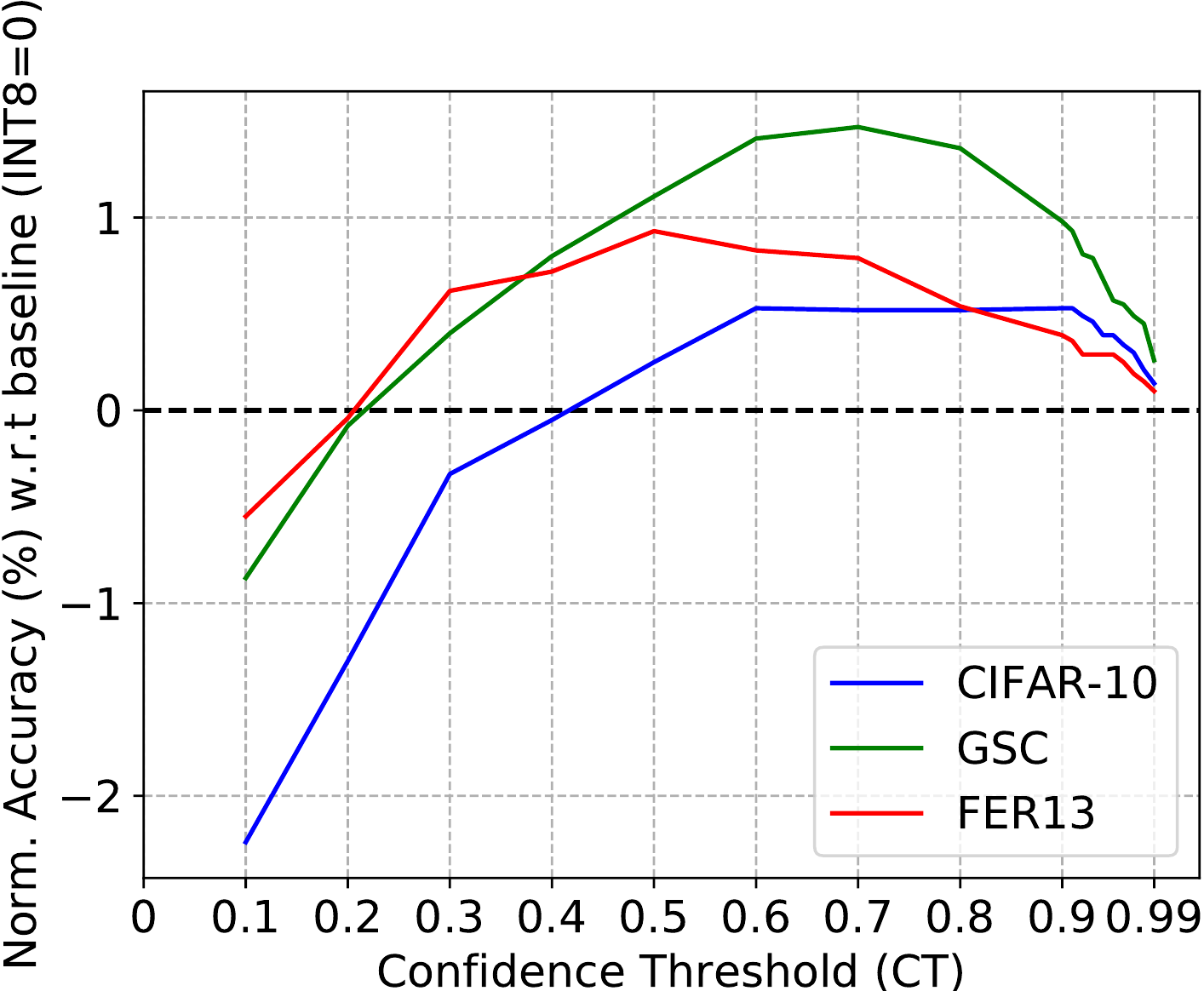}\vspace*{-1mm}
    \caption{ Normalized Accuracy (\%) w.r.t the INT8 baseline vs Confidence Threshold (CT)\vspace*{-1mm}}
    \label{fig:acc_cscore}
\end{figure}

The conducted experiments aimed at assessing the latency-accuracy trade-off. With this purpose, we first provide a parametric analysis that leverages the confidential threshold $CT$ as main knob.
The line plot in Fig. \ref{fig:acc_cscore} shows the delta accuracy achieved by CoopNet using as ground the accuracy of the baseline model INT8. The three tasks show the same trend: the CoopNet gets more accurate (positive delta) for larger values of $CT$. The break-even point $CT_{be}$ (for which delta is 0) may change depending on the complexity of the data-set and the classification capability of the CNN adopted: $CT_{be}=0.2$ for FER and GSC, $CT_{be}=0.4$ for CIFAR-10. To notice that the use of a confidential threshold $CT>CT_{be}$ guarantees substantial accuracy improvement. This suggests that CoopNet does not just improve over standard binarized CNNs, but it can also go beyond 8-bit quantization.
%However, it is important to emphasize that the accuracy gain comes at the cost of a reduced speed-up. 

\begin{figure}[!t]
    \centering
    \includegraphics[scale=0.43]{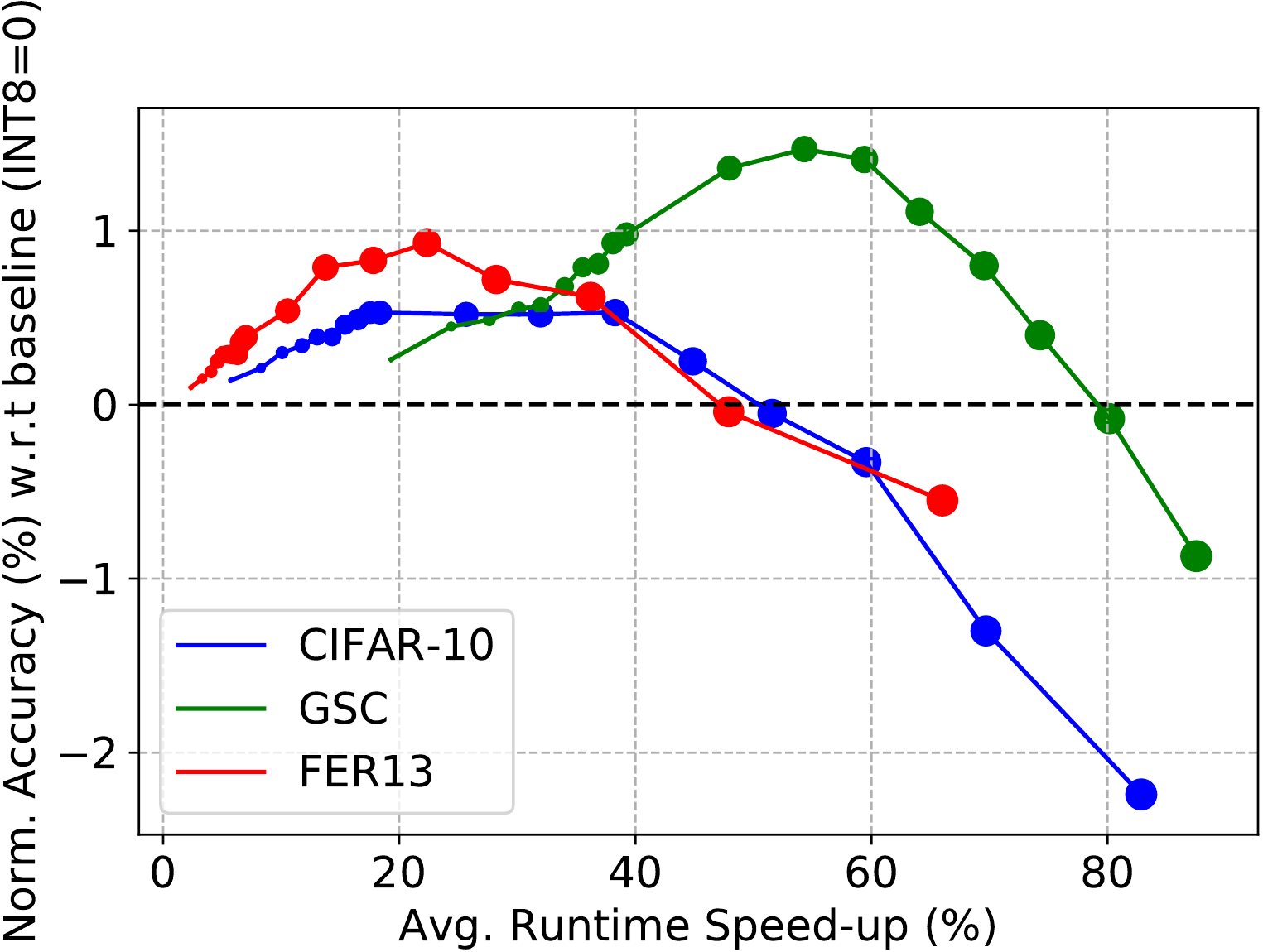}
    \caption{Normalized Accuracy (\%) and Average Speed-up w.r.t  baseline INT8 on test-set}
    \label{fig:acc_speedup}
\end{figure}

Even more interesting is the gain in terms of latency. Fig. \ref{fig:acc_speedup} shows the delta accuracy w.r.t. the baseline (INT8) as function of the average speed-up measured over the test set. The colored bullets drawn over the lines correspond to the actual value of $CT \in [0,1)$; to notice that bullets size is inversely proportional to the $CT$ value adopted: the smaller the $CT$, the larger the speed-up. Indeed, a large $CT$ implies that the BNN results get accepted as gold even if highly uncertain, hence the INT8 computation is skipped for speed-up. At the break-even, i.e. $CT=CT_{be}$, CoopNet shows impressive performance boost: 47.90\% for FER, 51.58\% for CIFAR-10, 80.16\% for GSC.
%That parameter affects both accuracy and speed-up: a conservative choice (higher ST) drastically reduces the speed-up gain margin while a lower ST implies a significant accuracy drop.

\begin{comment}
\begin{figure}[ht]
    \centering
    \includegraphics[scale=0.4]{img/accuracy_speedup_all.pdf}
    \caption{Accuracy w.r.t baseline vs Average Speed-up}
    \label{fig:acc_speedup}
\end{figure}
\end{comment}

Table \ref{tab:fin_res} gives a summary of some key results achieved by CoopNet. More specifically, it shows the evaluated extra-functional metrics (Speed-up and RAM footprint) under three accuracy level scenarios: $(i)$ CoopNet meets the accuracy of FP32, $(ii)$ CoopNet meets the accuracy of INT8 (the ground, i.e. $\Delta=0$), $(iii)$ CoopNet with the highest accuracy. As one can see, CoopNet guarantees substantial gains even under very high accuracy constraints. For instance, for GSC it achieves the same accuracy of the FP32 model with an average speed-up of 69.53\%; more interesting CoopNet can even overtake the FP32 model (+1.47\%). 
%This aspect is peculiar: that emerges from the joint combination of different CNNs that miss-classify different inputs. 
We observed that the joint action of BNN and INT8 helps to recognize inputs for which the FP32 model fails.
%This is not related to the arithmetic precision, but rather it depends on the topology of the CNN model. 

%the INT8 accuracy with a speed-up of 58\% and the FP32 level with 50\% over the three benchmarks (on average). However, we demonstrate, as a remarkable result, that CoopNet can even overtake the FP32 accuracy of 0.94\% with a 33.63\% of speed-up (on average).